\documentclass[preprint]{elsarticle}

\usepackage{amssymb}

\usepackage{natbib}

\usepackage{booktabs}       
\usepackage{graphics}        
\usepackage{caption}			
\usepackage{subfig}			

\usepackage{etoolbox}
\makeatletter
\patchcmd{\ps@pprintTitle}{\footnotesize\itshape
       Preprint submitted to \ifx\@journal\@empty Elsevier
       \else\@journal\fi\hfill\today}{\relax}{}{}
\makeatother

\begin{document}

\begin{frontmatter}



\title{Convolutional Neural Networks for the segmentation of microcalcification in Mammography Imaging}


\author[1,2]{Gabriele Valvano\corref{cor1}}
          \cortext[cor1]{Corresponding author.} 			
          \ead{gabriele.valvano@imtlucca.it}
\author[2]{Gianmarco Santini}
\author[2]{Nicola Martini}
\author[2]{Andrea Ripoli}
\author[3]{Chiara Iacconi}
\author[2]{Dante Chiappino}
\author[2]{and Daniele Della Latta}

\address[1]{IMT School for Advanced Studies Lucca, Lucca, Italy}
\address[2]{Imaging department, Fondazione Gabriele Monasterio, Massa, Italy}
\address[3]{Azienda USL Toscana Nord Ovest (ATNO), Carrara, Italy}

\begin{abstract}
Cluster of microcalcifications can be an early sign of breast cancer. In this paper we propose a novel approach based on convolutional neural networks for the detection and segmentation of microcalcification clusters. In this work we used 283 mammograms to train and validate our model, obtaining an accuracy of 98.22\% in the detection of preliminary suspect regions and of 97.47\% in the segmentation task. Our results show how deep learning could be an effective tool to effectively support radiologists during mammograms examination.
\end{abstract}

\begin{keyword}
Convolutional neural network \sep deep learning \sep segmentation \sep microcalcification \sep mammography imaging


\end{keyword}

\end{frontmatter}


\section{Introduction}
\label{sec_intro}

Breast cancer is one of the most common malignant neoplasm in the female population. The referral examination used for screening of breast cancer is Mammography.

Mammography is a radiological procedure that uses a bundle of X photons to map the breast tissue attenuation. With the use of high-resolution detectors, it is possible to detect microstructures with a high atomic number in the breast. Among them breast microcalcification (MC) can be an indicator for the diagnosis of breast cancer as they are the expression of cell necrosis.

In mammograms, microcalcifications appear as regions with high intensity compared to the local background and they can vary in size and have shapes ranging from circular geometries to strongly irregular ones with sharp or soft contours.

The Breast Imaging Reporting and Data System (BIRADS) standardized the interpretation of MCs by defining a scale ranging from 2 (benign finding) to 5 (highly suspicious of malignancy) based on their shape, density and distribution within the breast.

An important type of benign calcification that can be seen incidentally on mammography are breast arterial calcifications (BAC), which seem to  correlate with coronary calcification. 
Breast vascular calcifications are differentiated from malignant and ductal calcifications by size, morphology, and distribution and appear as linear "tram tracks" (\citet{lai2012linear}) of calcification  along arterial walls with a winding rather than branching course on mammography.

Since there are studies (see \citet{ryan2017breast}) correlating the estimation of the patient cardiovascular disease (CVD) with the amount of calcium residing at vascular level inside the breast, then the exact identification of the pixels belonging to a calcification can become crucial to assess possible outcomes for the future (\citet{molloi2009reproducibility}). For this reason the proposed system not only localizes MCs inside the tissue, but also aims to provide the exact segmentation of these lesions.
					
Because of the variability of connective, glandular and adipose tissue within the breast, microcalcifications are often difficult to find even for experienced operators. The heterogeneity of the breast tissue and projective image capture geometry implicates the impossibility to use a simple density threshold to automatically detect MCs. In addition, it is difficult to carry out the research by means of morphological filtering operations due to the large variability of their geometry.

In literature, a wide range of algorithms have been proposed for the automatic detection of clusters of mammary calcifications, highlighting the importance of this task. The first attempts were mainly based on the spatial characteristics of these lesions; an example of that is the morphological system proposed by \citet{Zhao1992}. Given the appearance of MCs as a locally high-intensity region, this work introduced a method based on the application of an adaptive thresholding operation on the mammogram, up to subsequently extract the lesions.

Subsequently, \citet{Wang1998} proposed an approach employing the wavelet transform to emphasize local variations in contrast. The intuition behind wavelets usage resides in their ability to discriminate different frequency bands and the possibility to preserve signal details at different resolutions. In this context microcalcifications correspond to high-frequency components in the image and they can be detected by decomposing the mammograms into different frequency subbands and filtering out low frequency variations from the image. 

Another proposal for the MCs detection pipeline is the multiresolutional analysis carried out by \citet{Netsch1999}, where the detection of microcalcifications is based on the Laplacian scale-space representation of a mammograms.

Later on, several papers proposed machine learning approaches to solve the task. Particularly, \citet{Edwards2000} formulated the MCs detection task as a supervised-learning problem and employed a Bayesian Neural Network to detect true MCs among several candidates obtained by a preliminary analysis of the mammogram. A second machine learning approach is the one proposed by \citet{El-Naqa2002}, who investigated instead, the possibility to apply support vector machines to develop the detection algorithm. 

Unfortunately, even if in some cases these methods were able to achieve a good sensitivity (i.e. \citet{El-Naqa2002} a sensitivity of 94\% was achieved, outperforming all the other methods tested by the authors), most of the previous approaches usually suffer from a high false-positive rate. This weakness is a direct consequence of the great variability of the breast tissue which must be taken into account to avoid misses of true positives also in very different mammograms. 

Given the difficulty which classical methods show in the detection problem, in this paper we propose the usage of a non-linear approach based on convolutional neural networks (CNNs). CNNs allow infact to avoid the direct definition of features to be analyzed in the image, providing both an automatic extraction and evaluation of these features in order to localize microcalcifications.

Going further, in order to accelerate the evaluation of mammographic images, we propose the usage of two CNNs to quickly detect the candidate region of interests (ROIs) and subsequently segment them. 

Therefore the final system should identify both the calcification clusters, for the subsequent detection of the presence of possible cancers, and the BAC for the CVD stratification. In this work, however, we focus only the segmentation of MCs and the cluster detection, without making any clinical assessment of patient risk. 

\section{Materials and Methods}
\label{sec_mm}
\subsection{Model}

Mammograms are high resolution images and they can correspond to big matrices (e.g. 4095x5625 pixels) which could be time-expensive to analyze. For this reason we developed a model consisting of two CNNs: we called the first CNN Detector while the second one was called Segmentator. Detector role is detecting candidate ROIs to be analyzed, while Segmentator classifies every pixel inside the given ROI.

The process of suspect ROIs detection must be non-computationally expensive because its role is to accelerate the processing of the whole mammographic image, then we pre-processed input images using Otsu thresholding to detect background pixels and exclude them from further evaluation.

We implemented both neural networks in Python, using the open-source software library Tensorflow.

We chose a patch-based approach to process the input mammograms. By using this kind of approach we assumed irrelevant to take into account the entire image for the classification of individual pixels, while we considered sufficient the local information. Moreover – by contrast with fully convolutional approaches – a patch-based approach allowed us to considerably increase the training set and easily perform a good data augmentation. 

With this purpose, we extracted squared patches with NxN dimension and the annotated labels from the available mammograms and segmentation masks. Labels for the Detector were chosen \textit{positive} if the relating patch contained a microcalcification inside, \textit{negative} in the opposite case. On the other hand, labels for Segmentator were assigned as \textit{positive} or \textit{negative}, accordingly to their central pixel: being part of a calcification or not.

In the proposed model, once completed the training process of the networks and during test time, the input mammogram is patched at run-time with NxN windows overlapped by N/2 pixels on each direction. The overlapping area leads to a redundancy able to limit misses of positive patches during the ROIs detection process. 

\begin{figure}[t]
      \centering
          \includegraphics[height=0.40\textwidth]{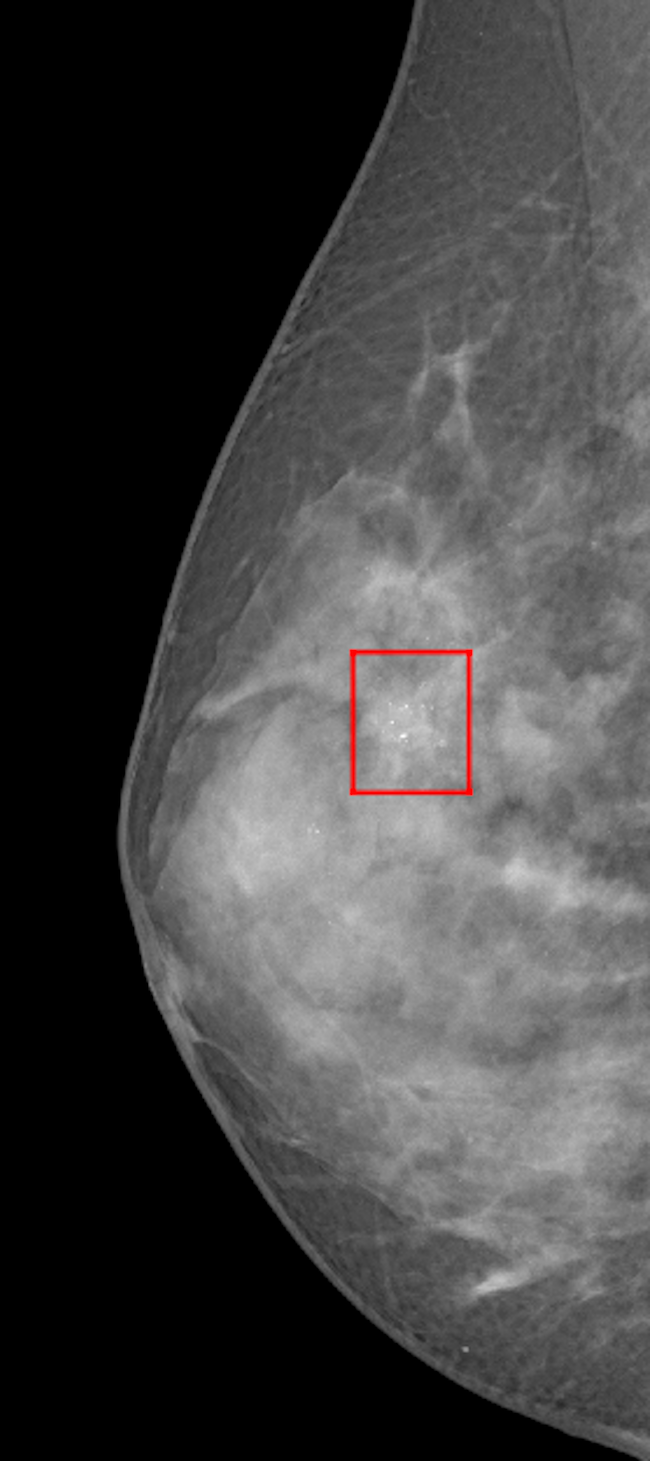}
          \hspace{0.5cm} 
          \includegraphics[height=0.40\textwidth]{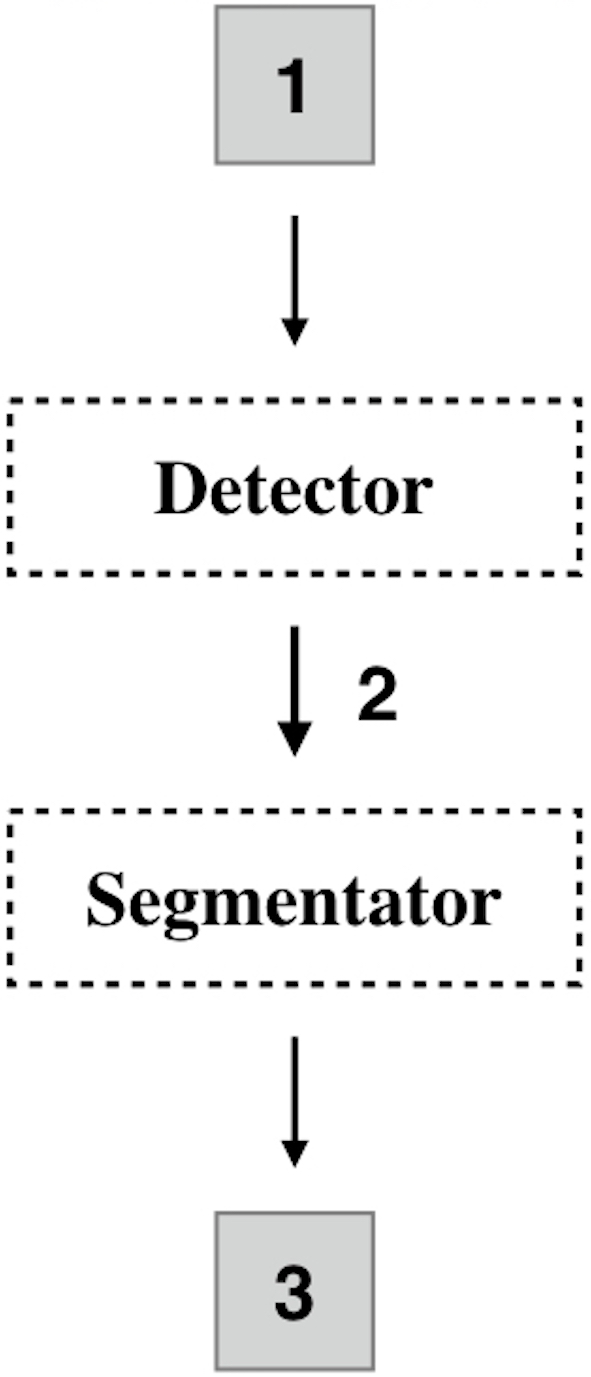}
          \vspace{0.2cm} 
          \\
          \includegraphics[width=0.17\textwidth]{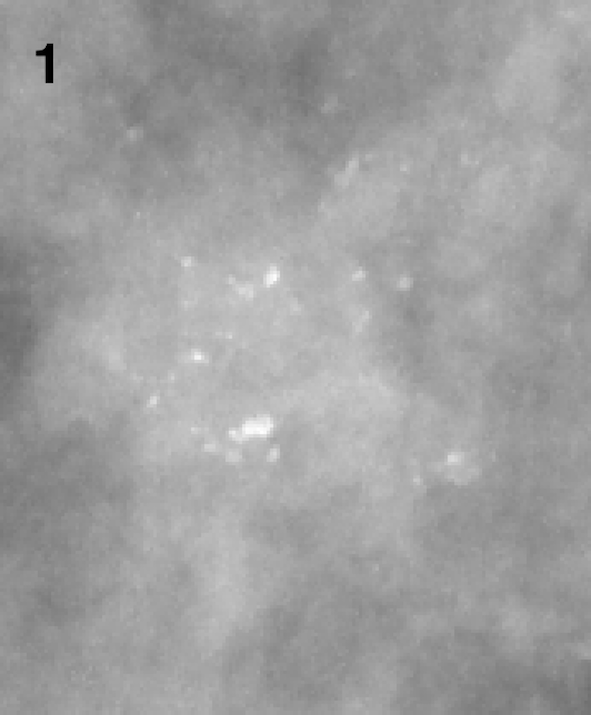} \hspace{0.05cm} 
          \includegraphics[width=0.17\textwidth]{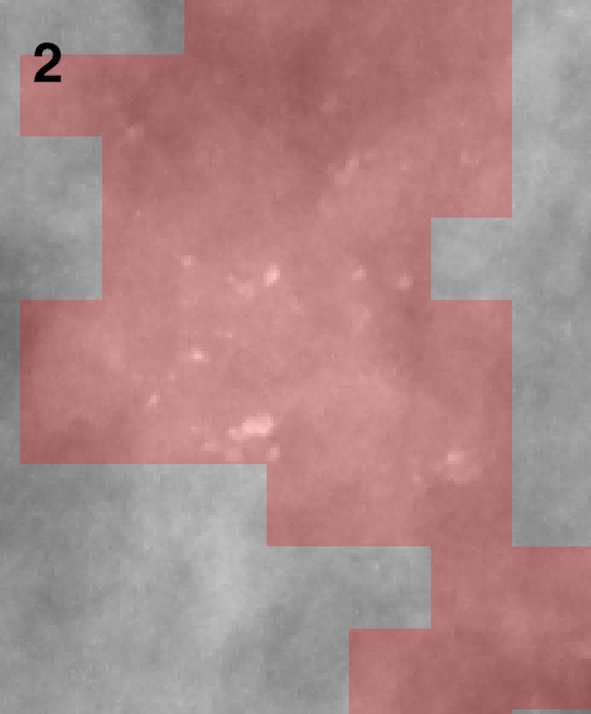} \hspace{0.05cm} 
          \includegraphics[width=0.17\textwidth]{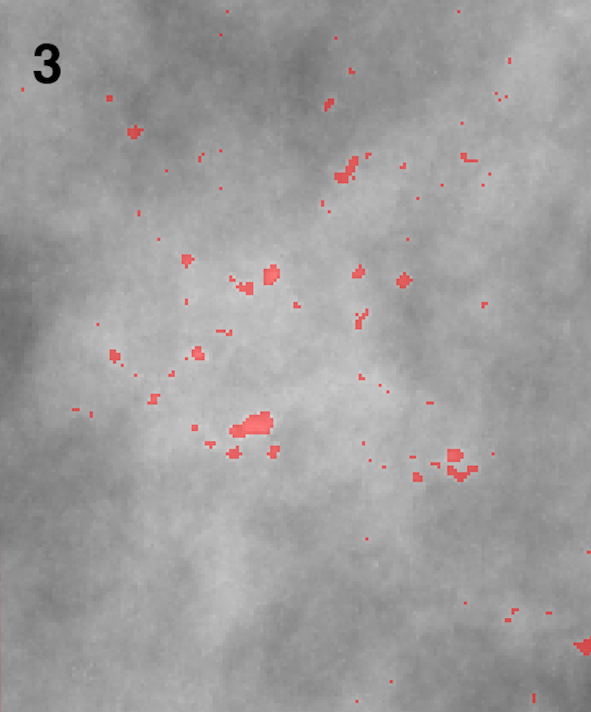}
          
      \caption{Example of breast segmentation process on a zoomed region in the mammogram. Best viewed in color.}
      \label{fig_segmentation_process}
\end{figure}

Once that the image has been patched, the Detector classifies each of these NxN input with a binary label relating to the potential presence of MCs inside. Candidate ROIs are subsequently segmented by Segmentator CNN, which brings to the creation of a binary mask of MCs (Figure \ref{fig_segmentation_process}).

The resulting mask is then analyzed via a labeling algorithm in order to localize the position of every MC inside the matrix. Clusters are considered inside regions with more than 5 distinct MCs on cm$^2$, as the radiological definition suggests (\citet{park2000clustering}).

\subsection{Data}
For our experiments we used 283 mammography images with a resolution of 0.05 mm. Among these images there are both natively digital mammograms and digitized images. 

Every image is associated to the corresponding manual segmentation mask realized by a breast imaging radiologist. Since each segmentation mask consisted in a binary matrix, classifying every pixel as a part of a MC or otherwise, we could utilize them to validate the obtained results.

In order to train our neural networks, the binary masks were also utilized to create the labels to be fed together with training samples to the CNNs, as ground truth.

We randomly chose 231 mammograms and the annotated labels to build the training set while 25 mammographic images were used to validate intermediate results and compare different networks architectures. The remaining 27 images were taken apart to build the test set and measure the final performances.

\subsection{Construction of training, validation and test set}
We experimented different values of patch dimension N. The final dimension was chosen as a compromise between giving enough information as CNN input and the need of maintaining low the computational burden of the model. 

\begin{figure}[t]
      \centering
          \includegraphics[width=0.4\textwidth]{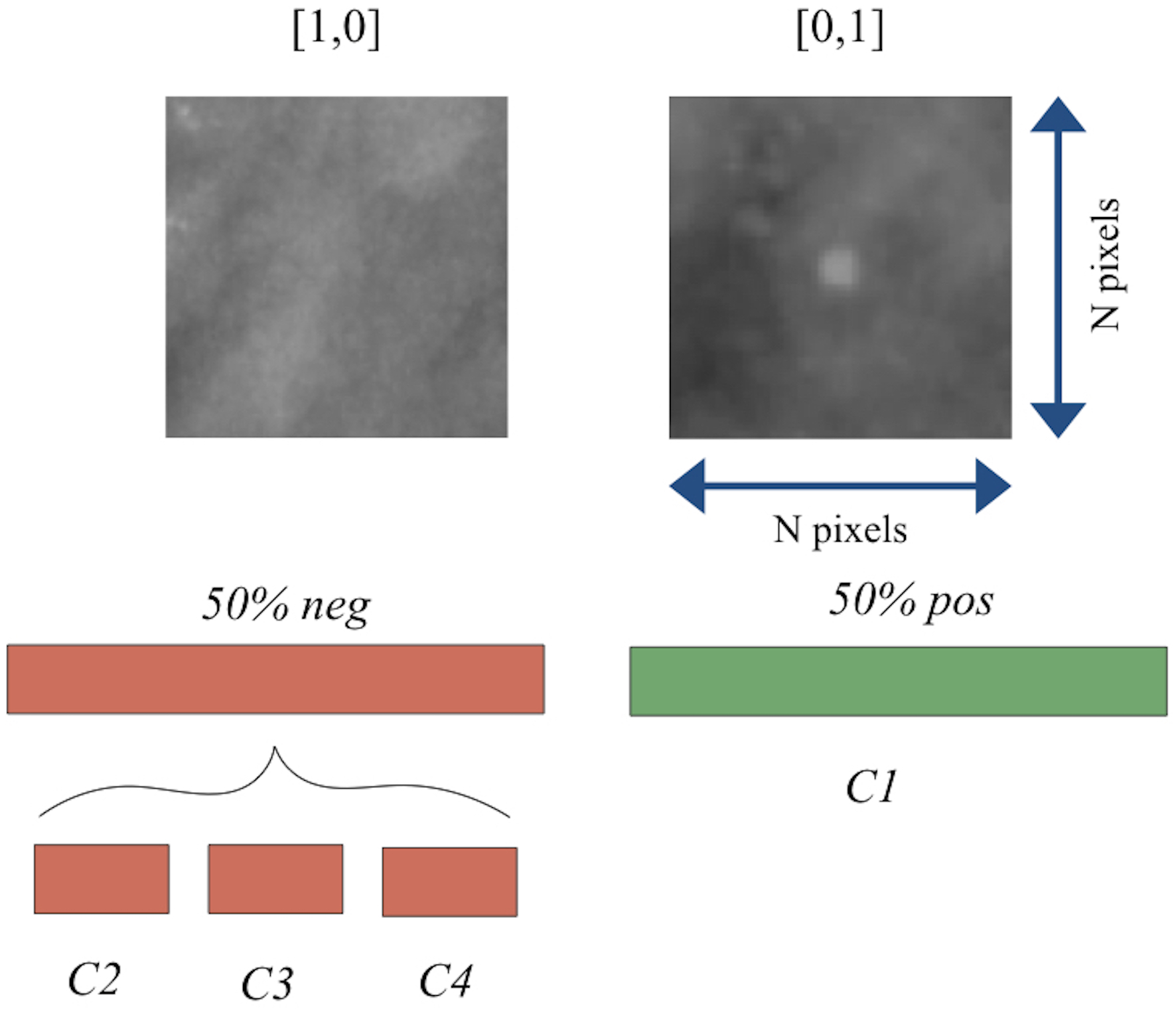}
      \caption{Example of patches and their subdivision in 4 different classes.}
      \label{fig_balanced_patches}
\end{figure}

We paid particular attention to collect samples for training, validation and test set, trying to make sure that the networks could see always as many input typologies as possible. For this reason we contemplated 4 possible classes of patch (Figure \ref{fig_balanced_patches}) listed below:

\begin{itemize}
\item class C1: patches whose central pixel belongs to a microcalcification;
\item class C2: patches with MCs close to the center but with the central pixel not belonging to a calcification;
\item class C3: cases where a calcification resides inside the patch, but is located peripherally and the central pixel does not belong to a MC;
\item class C4: cases where no MC is present inside the patch.
\end{itemize}

Since MCs are small circumscribed regions inside mammograms, certainly class C4 contains the largest number of patches inside the database and class C1 is the less numerous class. 

Moreover, we considered patches of class C2 as those containing calcifications in a range of 2 to 3 pixels from the center. This is a tricky class because as consequence of partial volume effect MCs’ border is frequently weakly defined and the classification of these pixels is often uncertain.

We organized the training set in a SQLite database to gather a customizable access to its samples during training. In particular, during the training process we sampled patches belonging to each of these classes paying attention to feed the network with the same number of positive and negative samples, which means a good balance of input minibatches. Accordingly to \citet{santini2017automatic} we built each minibatch on the fly with random patches sampled from the database. Since this approach leads to always different minibatches $-$ as a limit case we could say we won't ever have exactly the same samples in two different minibatches $-$ we believe it could improve the regularizing effect of batch normalization because it adds more randomicity to mean and variance inside minibatches. Moreover we found this strategy useful to train networks with such a strongly unbalanced dataset.

In addition, we used data augmentation during training-time to enlarge dataset dimension with artificial samples, obtained randomly rotating and flipping the images.

Patches for the validation set and test set were extracted from the 52 mammograms excluded from the training. Even in those cases each set contained a balanced number of samples, considering the presence of each class inside. 

\subsection{Networks architecture}
\begin{figure*}[ht]
      \centering
          \includegraphics[width=1\textwidth]{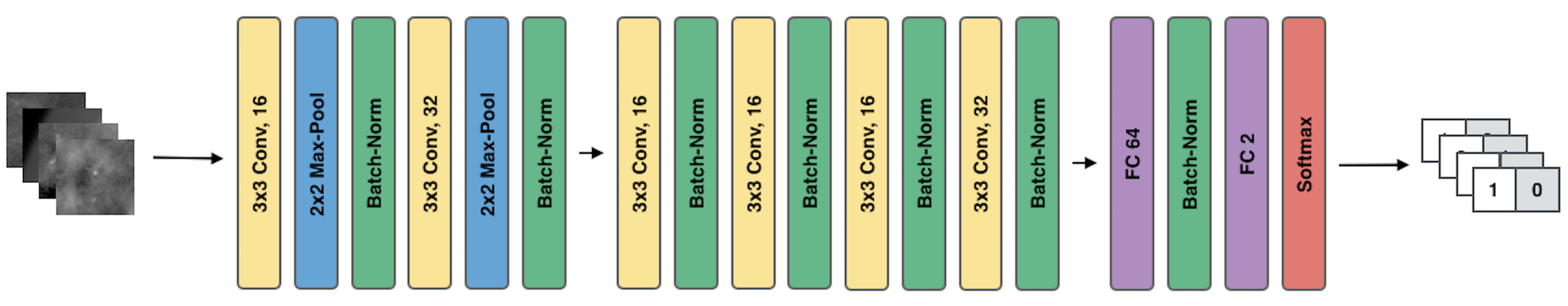}
      \caption{Neural network architecture implemented for Segmentator and Detector.}
      \label{fig_cnn_architecture}
\end{figure*}

Both Detector and Segmentator CNN share the same architecture (Figure \ref{fig_cnn_architecture}) consisting of 6 convolutional layers using 3x3 kernels and stride 1. 

In particular, we tested the difference between the usage of a \textit{same} convolution and a \textit{valid} convolution for both the Detector and the Segmentator architectures. This means that we tested the possibility of applying the zero padding operation throughout the network layers in order to preserve the input dimension unchanged while going deeper. This trial aimed to understand if the \textit{valid} convolution could invite the Segmentator to gradually reduce the input patch to features concerning only its central regions (i.e. from 49x49 patches to 2x2 features inputs to the final fully connected layers) and help it during the classification task; on the other hand, since the Detector should extract features from the whole input patch, we wondered if it could be favored by not reducing inputs through convolutions. In the results we demonstrate the differences between these two approaches.

In the architecture, the first and the second convolutional layers are followed by a max-pooling layer with 2x2 kernel to reduce the computational burden and induce the network to extract more abstract representations of the data. Two fully-connected layers of 64 and 2 hidden units close the architecture. We additionally used the drop-out strategy with 50\% probability over the 64-units fully-connected layer to limit network overfitting. 

Exception made for the last fully connected layer, each one of the above layers is followed by a batch normalization layer, which we found gave a consistent speed up in the learning process, as suggested by \citet{Ioffe2015}, by reducing internal covariate shift. Finally, we used He initialization (\citet{He2015delving}) to set the initial values for the weights and lessen their dependence on the initial state.

For each pixel $y_i$ of the vectorized output matrix $Y$ the posterior probability to belong to the \textit{l}-th class given the vectorized input patch $X$ and the network weight matrix $W$, is computed by the softmax classifier as:
\begin{eqnarray*}
p( y_i=l | X) = \frac {e^{s_l}} {\sum_{k=1}^K e^{s_k}}    \qquad     l = 1, ..., K
\end{eqnarray*}
where $s$ is the score function defined as $ s = f(X, W) $, given $f(\cdot)$ the non-linear function modelled by the neural net. With this formalism $s$ corresponds to the unnormalized log probabilities of the classes.

During the training we employed Stochastic Gradient Descent with Adam optimizer (\citet{kingma2014adam}) to minimize the Categorical Crossentropy: 
\begin{eqnarray*}
H( y, \widetilde{y} ) = -\sum_j y_j \cdot  log(\widetilde{y_j})
\end{eqnarray*}
where $y_j$ is the ground truth label for the \textit{j}-th class and $\widetilde{y}_j$ is the network output over that class. 

We trained each network with 256 sized mini-batches. Moreover, we chose a learning rate of $1\times10^{-3}$, but we manually halved this value when the loss plateaued in order to accelerate the convergence toward the minimum of the cost function. 

We applied early stopping strategy, taking the last obtained model configuration before any evidence of overfitting on training set.

\begin{table*}[t]
  \centering
	\resizebox{1\columnwidth}{!}{

  \begin{tabular}{llllll}  
    \multicolumn{6}{c}{Detector}                   \\
    \cmidrule{1-6}
         & Class C1   & Class C2 & Class C3 & Class C4 & Overall test accuracy\\
    \midrule
    		\textit{Valid}-architecture       & 0.19              &1.22       	& \textbf{20.28}         & 1.69         &  96.04				   \\
    		\textit{Same}-architecture     & 0.09              &0.47      	& \textbf{7.84}         & 2.09         &  \textbf{98.22}    	  \\
    \bottomrule
  \end{tabular} }

\vspace{0.3cm}

  \centering
  \resizebox{1\columnwidth}{!}{
  
  \begin{tabular}{llllll}  
    \multicolumn{6}{c}{Segmentator}                   \\
    \cmidrule{1-6}
         & Class C1   & Class C2 & Class C3 & Class C4 & Overall test accuracy\\
    \midrule
    		\textit{Valid}-architecture       & 1.85              & \textbf{15.32}          	& 0.34        & 0.55         & 96.37  				\\
    		\textit{Same}-architecture     & 1.57            	& \textbf{10.03}        &0.24  		 	& 0.20         & \textbf{97.47}   	\\
    \bottomrule
  \end{tabular} }
  
  \caption{Tables showing the obtained test error rate for each class and the overall test accuracy for the Detector CNN and the Segmentator CNN, using both \textit{valid} and \textit{same} convolution.}
  
  \label{tab_table_acc}
\end{table*}

\section{Results}
\label{sec_res}
Patch dimensions with side of 29, 39 and 49 pixels were tested. The best obtained results are a final accuracy of 98.22\% for Detector CNN and an accuracy of 97.47\% for Segmentator CNN. These results relate to the usage of patch dimension N=49, which gave better performance, and the employment of a convolution of type \textit{same}.

Table \ref{tab_table_acc} shows the error rate obtained for each contemplated class and the overall accuracy obtained on test samples. In particular, the table presents the values obtained using a \textit{valid} convolution approach and a \textit{same} convolution approach for both Detector and Segmentator.

\begin{figure*}
      \centering
      \large{Detector}\par
        \vspace{0.5cm}
        \includegraphics[width=1\textwidth]{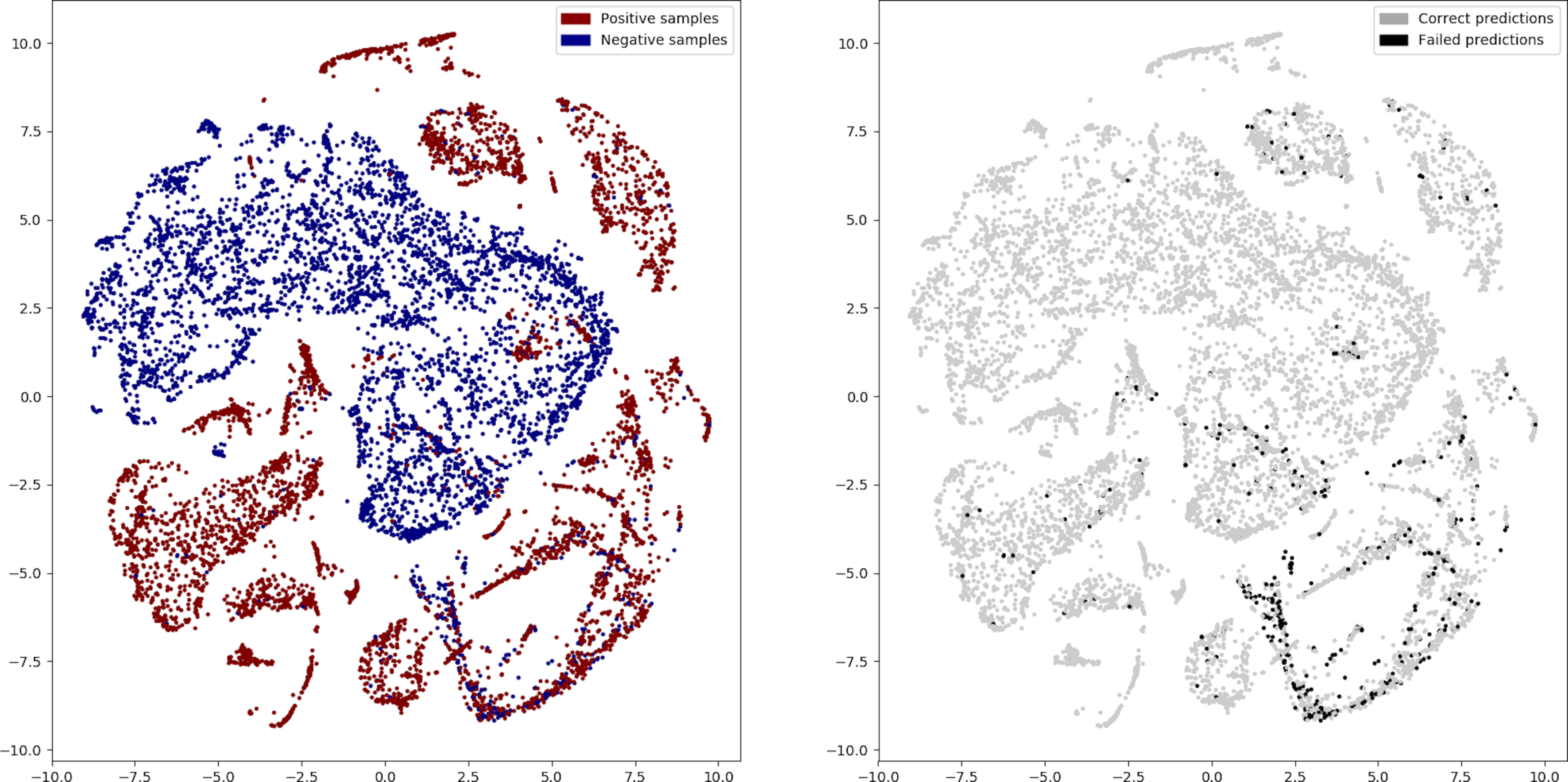} 
        \\
        \vspace{0.5cm}
        \large{Segmentator}\par
        \vspace{0.5cm}
       \includegraphics[width=1\textwidth]{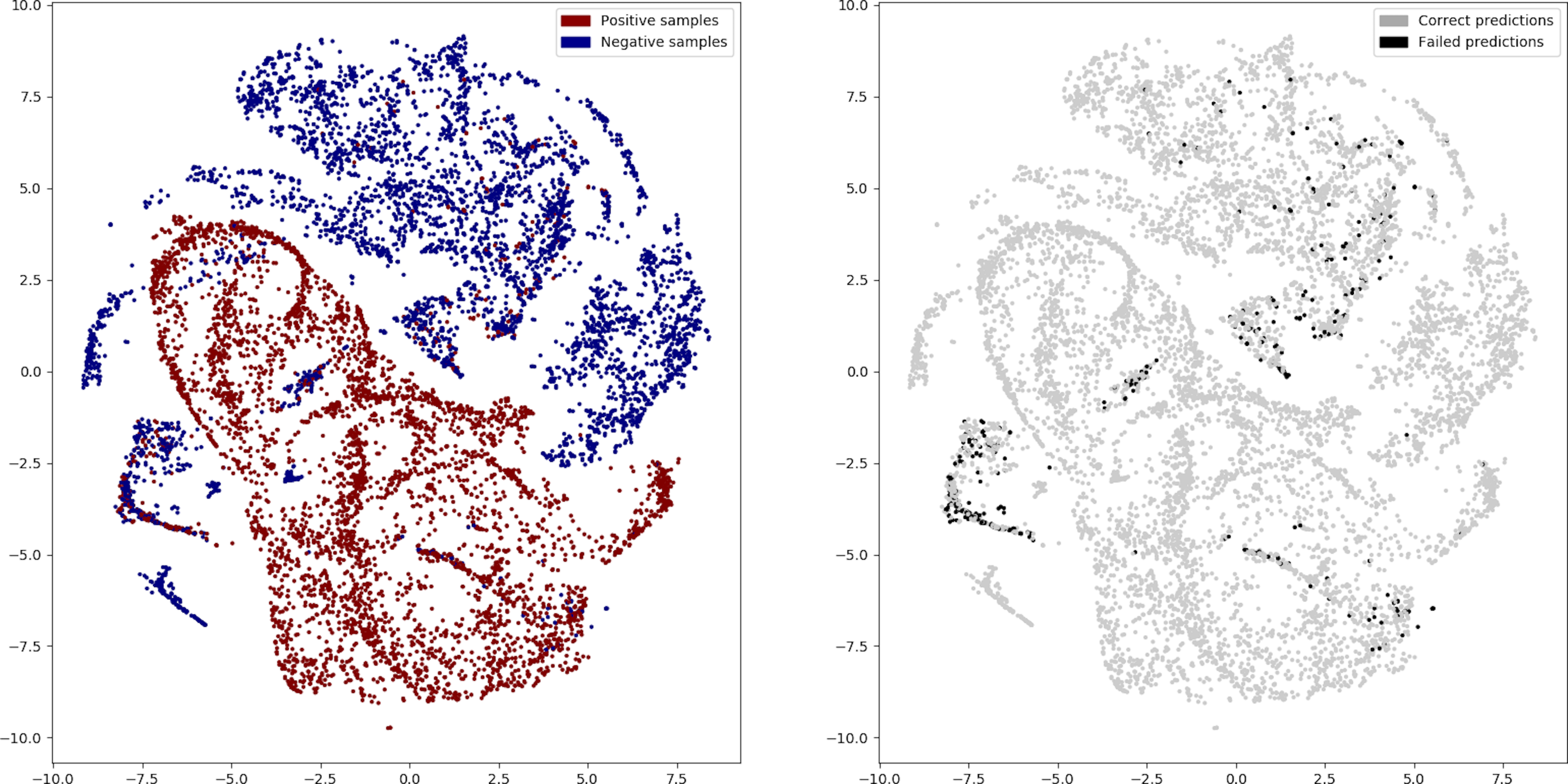} 
       \vspace{0.2cm} 
      \caption{Latent features spaces of the first fully-connected layer projected on 2D plane for the Detector and Segmentator neural networks. On the left: projection of samples from the \textit{positive} and \textit{negative} classes. On the right: misclassified samples position in the bidimensional projected spaces. Best viewed in color.} 
      \label{fig_tsne_2d}
\end{figure*}

\begin{figure*}
      \centering
      

			\includegraphics[width=0.63\textwidth]{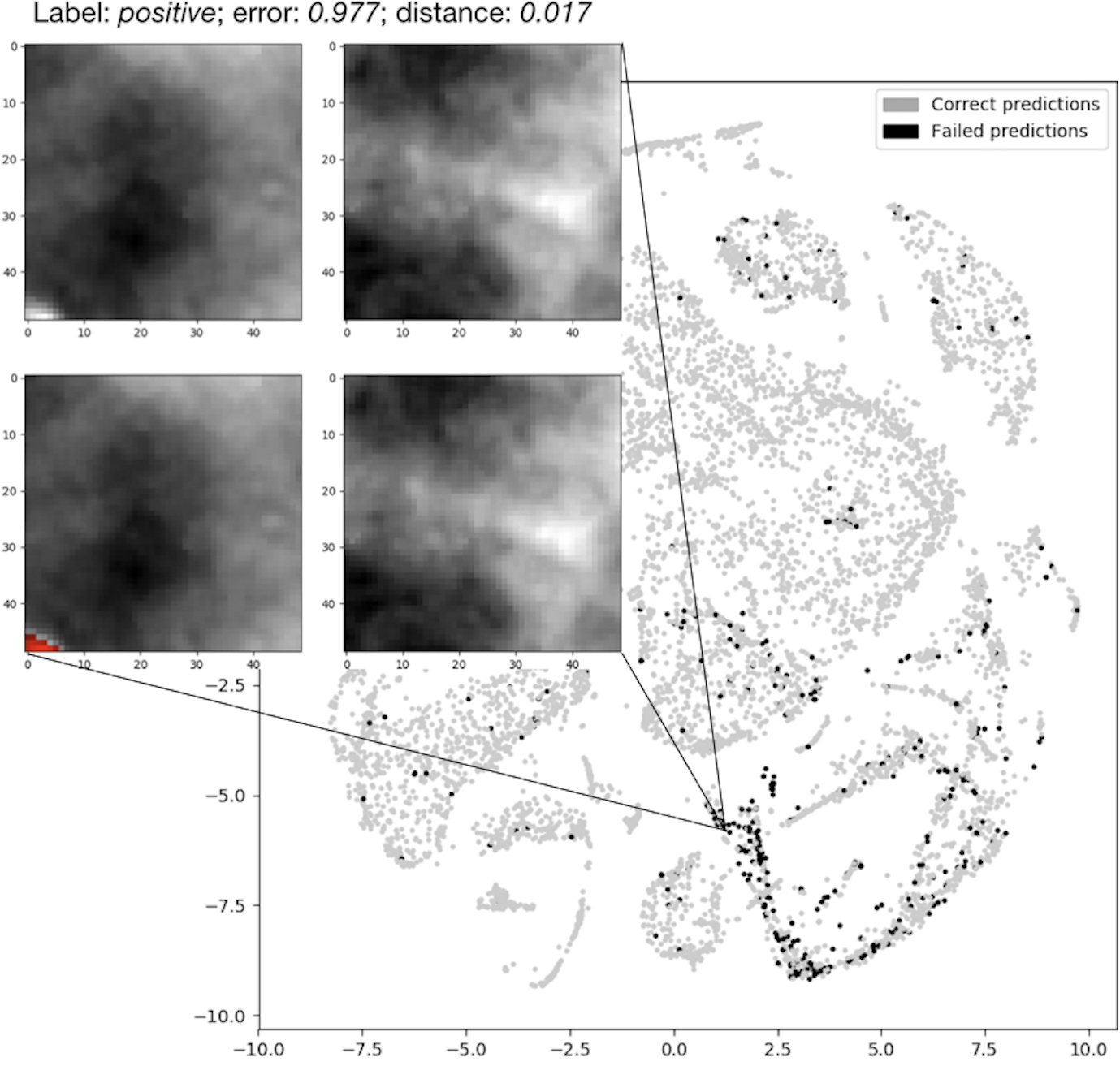}  
          
      \caption{Example of Detector failure case. The patch on the left represents the misclassified input sample containing a microcalcification, while the patch on the right is the $-$ well classified $-$ closest class C4 sample in the features space. Below, you can see the ground truth segmentations. Please note that the maximum possible error is equal to 1 and an error $<$ 0.5 means that the input patch is still correctly classified. Best viewed in color.}
	   \vspace{0.5cm} 
		
       \label{fig_failed_patches_1}

			\includegraphics[width=0.63\textwidth]{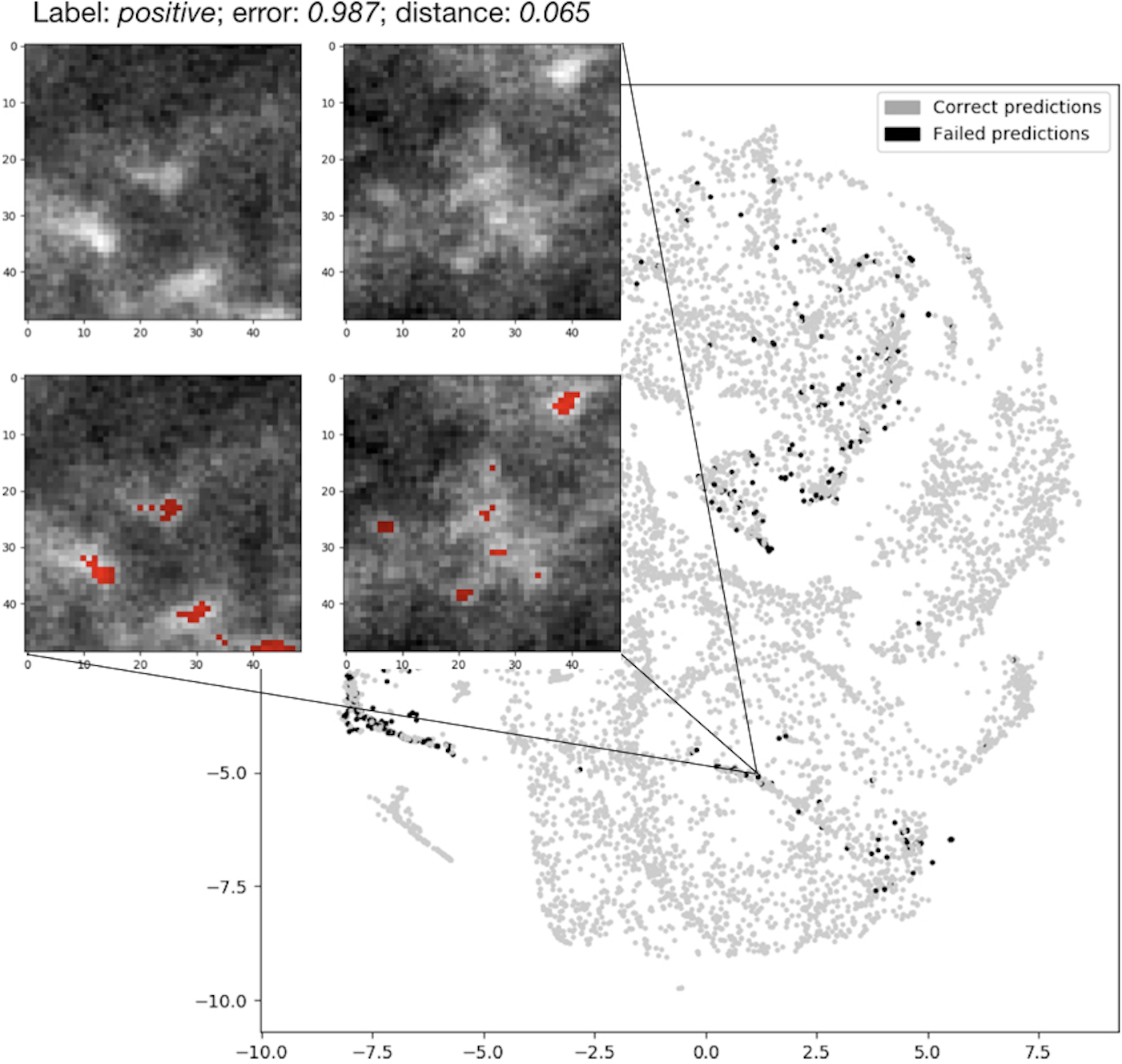}          
          
      \caption{Example of Segmentator failure case. The patch on the left represents the misclassified input sample containing a microcalcification, while the patch on the right is the $-$ well classified $-$ closest class C2 sample in the features space. Below, you can see the ground truth segmentations. Please note that the maximum possible error is equal to 1 and an error $<$ 0.5 means that the input patch is still correctly classified. Best viewed in color.}
      
       \label{fig_failed_patches_2}
\end{figure*}

To deepen our understanding of the network behavior we also conducted an analysis of misclassified patches in the features domain using t-SNE (\citet{maaten2008visualizing}). Figure \ref{fig_tsne_2d} shows the features spaces before the last fully connected layer projected on 2D planes for either Segmentator or Detector. Figure \ref{fig_failed_patches_1} and \ref{fig_failed_patches_2} show examples of misclassified patches and their nearest neighbors in the features domain.

We tested the entire model on GPU GTX 970 and every mammogram was processed in roughly 20 seconds.

An example of input region segmentated by the model is represented in Figure \ref{fig_segmentation_process}. 

\section{Discussion}
\label{sec_disc}
From the results illustrated in Table \ref{tab_table_acc} we could experiment how performing a convolution of type \textit{same} instead of \textit{valid} led to a relevant difference in both Detector and Segmentator performance over each class. 

We believe these differences to be independent from stochastic oscillations of the cost function during the network training. Instead, this improvement can most probably be explained by the observation that using a \textit{same} convolution leads to larger inputs to the first fully connected layer and consequently to a larger number of weights at this level.

We consequently prefer the usage of a \textit{same} convolution approach in both cases since it leads to better results and subsequently to a higher generalization capability of the model.

%
%

Table \ref{tab_table_acc} also outlines how the hardly classified classes mainly relate to limit cases between the assignment of \textit{positive} or \textit{negative} labels. We could assume class C2 to be a tricky class for the Segmentator because – out of the partial volume effect – MCs’ borders are often soft and hardly evaluated even by humans. On the other hand it can be observed how class C3 achieves the worse performance during the detection process, which is probably due to the presence of patches with only a small number of peripheral pixels belonging to a MC (i.e with only the one most external pixel in the corner of the patch) and, again, often belonging to MCs' border regions.


As a confirmation of all these hypothesis we conducted an analysis of the misclassified patches both for Detector and Segmentator neural networks. In particular we analyzed extracted data representations from the penultimate fully connected layer using t-SNE.

We examined C2, C3 and C4 top misclassified patches from the Segmentator and extracted the nearest neighbors in the features domain belonging to class C1 and vice versa. On the other hand we examined C1, C2 and C3 top misclassified samples from Detector and corresponding nearest neighbor samples in the features domain which belonged to class C4 and vice versa. 

An example of meaningful misclassified patches and their closest \textit{positive} or \textit{negative} samples in the latent space is illustrated in Figure \ref{fig_failed_patches_1} and \ref{fig_failed_patches_2}. Visual analysis of results seems to confirm our preliminary interpretation of the errors. 

By contrast, we highlight how Table \ref{tab_table_acc} also points out that the error rate is maintained low in non-limit cases, which is desirable. In fact, for the Segmentator CNN this means that even if the network fails classifying boundary pixels of a calcification, it generally recognize its presence, achieving a good lesion-wise accuracy when the lesion resides in the central region of the input patch. At the same time, the Detector shows a good lesion-wise accuracy when MCs are well visible in the input data, failing only with more peripheral lesions. In this context we considered an overlap of half patch along each direction during the mammogram analysis, up to prevent MCs misses during the detection process and consequently breaking down detector error rate on class C3 samples.

\begin{figure}
      \centering
          \includegraphics[width=0.20\textwidth]{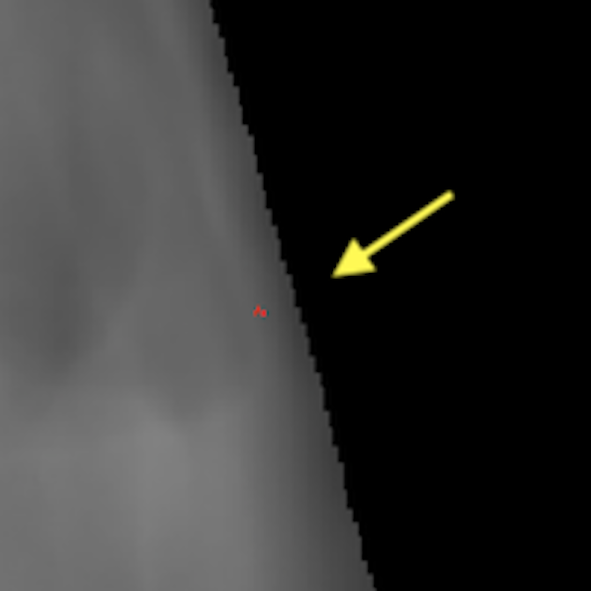} 
	  	 \hspace{0.01\textwidth}  
          \includegraphics[width=0.20\textwidth]{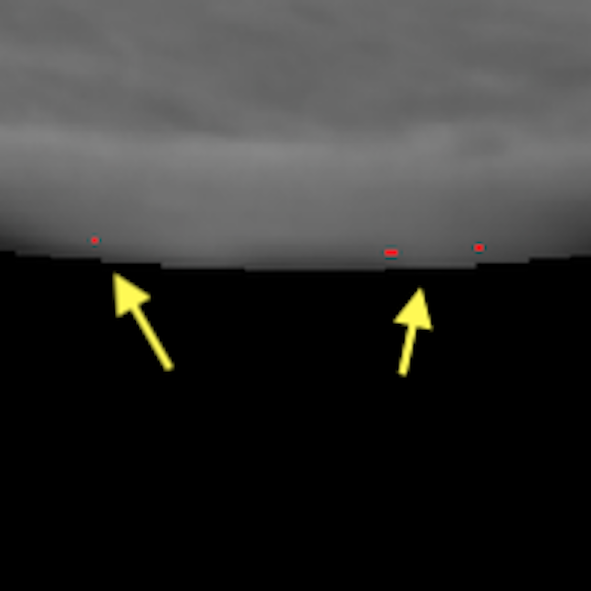}
      \caption{Example of false negatives in correspondence of the transition region from the breast tissue to the background pixels. Best viewed in color.}
       \label{fig_border_fails}
\end{figure}

In addition $-$ aside from tricky classes $-$ an interesting fact pointed out by a qualitative analysis of the segmentation masks concerns a certain inclination of the model to make mistakes in correspondence of the transition region from the breast tissue to the background pixels. This is probably due to the fact that these regions usually relate to areas with strong contrast variation. An example of the phenomenon is illustrated in Figure \ref{fig_border_fails}.

Finally, this analysis highlighted how the major source of false positives seems to reside in the digitized images, where the presence of a greater quantity of widespread noise leads the network to commit a greater number of errors.

\section{Conclusion}
\label{sec_concl}
We propose a model to detect and segment breast microcalcifications within mammographic images. This model is composed of two consecutive blocks based on convolutional neural networks: the Detector and the Segmentator. Thanks to the preliminary analysis carried out by the the first CNN the computational burden is considerably reduced and the total segmentation process does not becomes time consuming.

Moreover, the quality of the achieved results suggest the potentialities of this tool to effectively support radiologists during mammograms examination, bringing aid during the non-trivial evaluation of uncertain regions and reducing the diagnosys time. This could be especially useful in the screening setting, where the large number of examinations could reduce the attention of the reader, to support diagnosis or to narrow differential diagnosis.




\section*{Conflict of interest statement}
None declared.



\section*{References}
\bibliographystyle{elsarticle-num-names.bst} 
\bibliography{reference_db.bib}

\begin{thebibliography}{14}
\providecommand{\natexlab}[1]{#1}
\providecommand{\url}[1]{\texttt{#1}}
\providecommand{\urlprefix}{URL }
\expandafter\ifx\csname urlstyle\endcsname\relax
  \providecommand{\doi}[1]{doi:\discretionary{}{}{}#1}\else
  \providecommand{\doi}[1]{doi:\discretionary{}{}{}\begingroup
  \urlstyle{rm}\url{#1}\endgroup}\fi
\providecommand{\bibinfo}[2]{#2}

\bibitem[{Lai et~al.(2012)Lai, Slanetz, and Eisenberg}]{lai2012linear}
\bibinfo{author}{K.~C. Lai}, \bibinfo{author}{P.~J. Slanetz},
  \bibinfo{author}{R.~L. Eisenberg}, \bibinfo{title}{Linear breast
  calcifications}, \bibinfo{journal}{American Journal of Roentgenology}
  \bibinfo{volume}{199}~(\bibinfo{number}{2}) (\bibinfo{year}{2012})
  \bibinfo{pages}{W151--W157}.

\bibitem[{Ryan et~al.(2017)Ryan, Choi, Choi, and Lewis}]{ryan2017breast}
\bibinfo{author}{A.~J. Ryan}, \bibinfo{author}{A.~D. Choi},
  \bibinfo{author}{B.~G. Choi}, \bibinfo{author}{J.~F. Lewis},
  \bibinfo{title}{Breast arterial calcification association with coronary
  artery calcium scoring and implications for cardiovascular risk assessment in
  women}, \bibinfo{journal}{Clinical cardiology} .

\bibitem[{Molloi et~al.(2009)Molloi, Mehraien, Iribarren, Smith, Ducote, and
  Feig}]{molloi2009reproducibility}
\bibinfo{author}{S.~Molloi}, \bibinfo{author}{T.~Mehraien},
  \bibinfo{author}{C.~Iribarren}, \bibinfo{author}{C.~Smith},
  \bibinfo{author}{J.~L. Ducote}, \bibinfo{author}{S.~A. Feig},
  \bibinfo{title}{Reproducibility of breast arterial calcium mass
  quantification using digital mammography}, \bibinfo{journal}{Academic
  radiology} \bibinfo{volume}{16}~(\bibinfo{number}{3}) (\bibinfo{year}{2009})
  \bibinfo{pages}{275--282}.

\bibitem[{Zhao et~al.(1992)Zhao, Shridhar, and Daut}]{Zhao1992}
\bibinfo{author}{D.~Zhao}, \bibinfo{author}{M.~Shridhar},
  \bibinfo{author}{D.~Daut}, \bibinfo{title}{Morphology on detection of
  calcifications in mammograms}, in: \bibinfo{booktitle}{Acoustics, Speech, and
  Signal Processing, 1992. ICASSP-92., 1992 IEEE International Conference on},
  vol.~\bibinfo{volume}{3}, \bibinfo{organization}{IEEE},
  \bibinfo{pages}{129--132}, \bibinfo{year}{1992}.

\bibitem[{Wang and Karayiannis(1998)}]{Wang1998}
\bibinfo{author}{T.~C. Wang}, \bibinfo{author}{N.~B. Karayiannis},
  \bibinfo{title}{Detection of microcalcifications in digital mammograms using
  wavelets}, \bibinfo{journal}{IEEE transactions on medical imaging}
  \bibinfo{volume}{17}~(\bibinfo{number}{4}) (\bibinfo{year}{1998})
  \bibinfo{pages}{498--509}.

\bibitem[{Netsch and Peitgen(1999)}]{Netsch1999}
\bibinfo{author}{T.~Netsch}, \bibinfo{author}{H.-O. Peitgen},
  \bibinfo{title}{Scale-space signatures for the detection of clustered
  microcalcifications in digital mammograms}, \bibinfo{journal}{IEEE
  Transactions on medical imaging} \bibinfo{volume}{18}~(\bibinfo{number}{9})
  (\bibinfo{year}{1999}) \bibinfo{pages}{774--786}.

\bibitem[{Edwards et~al.(2000)Edwards, Kupinski, Nagel, Nishikawa, and
  Papaioannou}]{Edwards2000}
\bibinfo{author}{D.~Edwards}, \bibinfo{author}{M.~Kupinski},
  \bibinfo{author}{R.~Nagel}, \bibinfo{author}{R.~Nishikawa},
  \bibinfo{author}{J.~Papaioannou}, \bibinfo{title}{Using a Bayesian neural
  network to optimally eliminate false-positive microcalcification detections
  in a CAD scheme}, \bibinfo{journal}{Digital Mammography, Medical Physics
  Publishing, Madison}  (\bibinfo{year}{2000}) \bibinfo{pages}{168--173}.

\bibitem[{El-Naqa et~al.(2002)El-Naqa, Yang, Wernick, Galatsanos, and
  Nishikawa}]{El-Naqa2002}
\bibinfo{author}{I.~El-Naqa}, \bibinfo{author}{Y.~Yang}, \bibinfo{author}{M.~N.
  Wernick}, \bibinfo{author}{N.~P. Galatsanos}, \bibinfo{author}{R.~M.
  Nishikawa}, \bibinfo{title}{A support vector machine approach for detection
  of microcalcifications}, \bibinfo{journal}{IEEE transactions on medical
  imaging} \bibinfo{volume}{21}~(\bibinfo{number}{12}) (\bibinfo{year}{2002})
  \bibinfo{pages}{1552--1563}.

\bibitem[{Park et~al.(2000)Park, Choi, Bae, Lee, Ahn, and
  Gong}]{park2000clustering}
\bibinfo{author}{J.~Park}, \bibinfo{author}{H.~Choi}, \bibinfo{author}{S.-J.
  Bae}, \bibinfo{author}{M.-S. Lee}, \bibinfo{author}{S.-H. Ahn},
  \bibinfo{author}{G.~Gong}, \bibinfo{title}{Clustering of breast
  microcalcifications: revisited}, \bibinfo{journal}{Clinical radiology}
  \bibinfo{volume}{55}~(\bibinfo{number}{2}) (\bibinfo{year}{2000})
  \bibinfo{pages}{114--118}.

\bibitem[{Santini et~al.(2017)Santini, Della~Latta, Martini, Valvano, Gori,
  Ripoli, Susini, Landini, and Chiappino}]{santini2017automatic}
\bibinfo{author}{G.~Santini}, \bibinfo{author}{D.~Della~Latta},
  \bibinfo{author}{N.~Martini}, \bibinfo{author}{G.~Valvano},
  \bibinfo{author}{A.~Gori}, \bibinfo{author}{A.~Ripoli},
  \bibinfo{author}{C.~L. Susini}, \bibinfo{author}{L.~Landini},
  \bibinfo{author}{D.~Chiappino}, \bibinfo{title}{An automatic deep learning
  approach for coronary artery calcium segmentation}, in:
  \bibinfo{booktitle}{EMBEC \& NBC 2017}, \bibinfo{publisher}{Springer},
  \bibinfo{pages}{374--377}, \bibinfo{year}{2017}.

\bibitem[{Ioffe and Szegedy(2015)}]{Ioffe2015}
\bibinfo{author}{S.~Ioffe}, \bibinfo{author}{C.~Szegedy}, \bibinfo{title}{Batch
  Normalization: Accelerating Deep Network Training by Reducing Internal
  Covariate Shift}, \bibinfo{journal}{arXiv preprint arXiv:1502.03167} .

\bibitem[{He et~al.(2015)He, Zhang, Ren, and Sun}]{He2015delving}
\bibinfo{author}{K.~He}, \bibinfo{author}{X.~Zhang}, \bibinfo{author}{S.~Ren},
  \bibinfo{author}{J.~Sun}, \bibinfo{title}{Delving deep into rectifiers:
  Surpassing human-level performance on imagenet classification}, in:
  \bibinfo{booktitle}{Proceedings of the IEEE international conference on
  computer vision}, \bibinfo{pages}{1026--1034}, \bibinfo{year}{2015}.

\bibitem[{Kingma and Ba(2014)}]{kingma2014adam}
\bibinfo{author}{D.~Kingma}, \bibinfo{author}{J.~Ba}, \bibinfo{title}{Adam: A
  method for stochastic optimization}, \bibinfo{journal}{arXiv preprint
  arXiv:1412.6980} .

\bibitem[{Maaten and Hinton(2008)}]{maaten2008visualizing}
\bibinfo{author}{L.~v.~d. Maaten}, \bibinfo{author}{G.~Hinton},
  \bibinfo{title}{Visualizing data using t-SNE}, \bibinfo{journal}{Journal of
  Machine Learning Research} \bibinfo{volume}{9}~(\bibinfo{number}{Nov})
  (\bibinfo{year}{2008}) \bibinfo{pages}{2579--2605}.

\end{thebibliography}

\end{document}